\documentclass[runningheads]{llncs}

\usepackage{eccv}

\usepackage{eccvabbrv}

\usepackage{graphicx}
\usepackage{booktabs}

\usepackage{enumitem}
\usepackage{bm}
\usepackage{multirow}

\usepackage{amsmath}
\usepackage{amssymb}
\usepackage{mathtools}
\usepackage[accsupp]{axessibility}  

\usepackage{hyperref}

\usepackage{orcidlink}

\makeatletter
\def\thanks#1{\protected@xdef\@thanks{\@thanks
        \protect\footnotetext{#1}}}
\makeatother

\begin{document}

\title{MinD-3D: Reconstruct High-quality 3D objects in Human Brain} 

\titlerunning{MinD-3D: Reconstruct High-quality 3D objects in Human Brain}

\author{
Jianxiong Gao\and Yuqian Fu$^\S$\and Yun Wang\and Xuelin Qian$^\ddag$ \\
Jianfeng Feng\and Yanwei Fu$^\dagger$
\thanks{$\S$: Dr. Yuqian Fu is now with ETH Zürich and INSAIT.}
\thanks{$\ddag$: Dr. Xuelin Qian is now with Northwestern Polytechnical University.}
\thanks{$^\dagger$: Corresponding author.}
}

\authorrunning{Gao et al.}

\institute{Fudan University\\
\email{jxgao22@m.fudan.edu.cn, \{fuyq20,19110850009,xlqian,jffeng,yanweifu\}@fudan.edu.cn}
}

\maketitle

\begin{abstract}
In this paper, we introduce \textbf{Recon3DMind}, an innovative task aimed at reconstructing 3D visuals from Functional Magnetic Resonance Imaging (fMRI) signals, marking a significant advancement in the fields of cognitive neuroscience and computer vision. To support this pioneering task, we present the \textbf{fMRI-Shape} dataset, which includes data from 14 participants and features 360-degree videos of 3D objects to enable comprehensive fMRI signal capture across various settings, thereby laying a foundation for future research. Furthermore, we propose \textbf{MinD-3D}, a novel and effective three-stage framework specifically designed to decode the brain's 3D visual information from fMRI signals, demonstrating the feasibility of this challenging task. The framework begins by extracting and aggregating features from fMRI frames through a neuro-fusion encoder, subsequently employs a feature bridge diffusion model to generate visual features, and ultimately recovers the 3D object via a generative transformer decoder. We assess the performance of MinD-3D using a suite of semantic and structural metrics and analyze the correlation between the features extracted by our model and the visual regions of interest (ROIs) in fMRI signals. Our findings indicate that MinD-3D not only reconstructs 3D objects with high semantic relevance and spatial similarity but also significantly enhances our understanding of the human brain's capabilities in processing 3D visual information. Project page at: \url{https://jianxgao.github.io/MinD-3D}.

  \keywords{fMRI \and 3D vision \and diffusion model}
\end{abstract}

\section{Introduction}
\label{sec:intro}

Functional Magnetic Resonance Imaging (fMRI), a kind of signal that can be obtained in a non-invasive way, could capture blood changes in the human brain induced by neuronal activity. Due to its relatively easy accessibility, fMRI has been commonly used to reflect visual activities.
Some recent studies~\cite{chen2023seeing,chen2023cinematic,scotti2023reconstructing,qian2023fmri,qian2023semantic} have successfully reconstructed high-quality images from fMRI signals by utilizing powerful generative models~\cite{rombach2021highresolution,chang2022maskgit}.
These approaches focus on extracting semantic features from fMRI signals, often requiring only semantic features to generate relevant high-quality images.

\begin{figure}[t]
    \centering
    \includegraphics[width=\columnwidth]{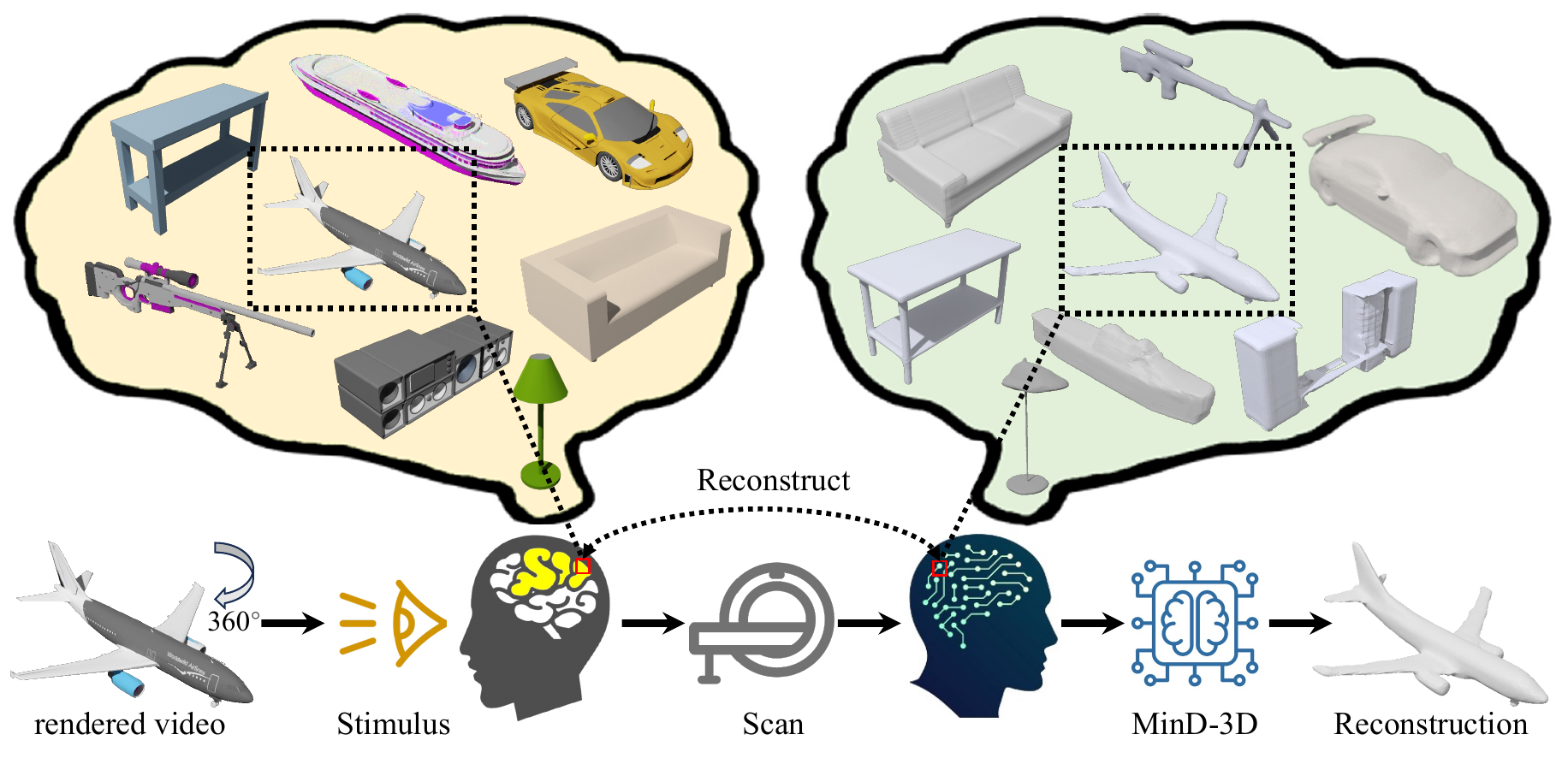}
\captionof{figure}{\label{fig:teaser}
Overview of Recon3DMind task, showcasing the fMRI-Shape dataset collection process with 14 participants observing 360-degree view videos of 3D objects, and MinD-3D framework for reconstructing 3D objects from fMRI signals.
}
\end{figure}

However, the human visual system encompasses more than the mere processing of 2D images; it possesses the remarkable ability to translate 2D projections into rich, 3D representations. 
This complex mechanism bestows upon us a rich perception of our world, complete with discernible aspects of size, distance, and depth.
Diverging from earlier studies, our research focuses on modeling this 3D visual capability of the brain. 
We introduce a novel task, termed \textbf{Recon}structing \textbf{3D} Objects from \textbf{Mind} in fMRI (\textbf{Recon3DMind}), that leverages advanced computer vision techniques to decode and reconstruct the 3D visual information perceived by the brain from fMRI signals. This task goes beyond extracting semantic features and includes spatial and structural dimensions, which are crucial for a thorough understanding of 3D visuals.

Some literature~\cite{groen2019scenes,10.1093/oso/9780190070557.003.0011,linton2023minimal} has shown that the brain's visual processing mechanisms for 3D perception are significantly more complex than those for 2D perception. This increased complexity is manifested in the differential activation of brain regions during 3D visualization tasks~\cite{georgieva2009processing,jerath2015functional}. 
Consequently, a reliance solely on semantic features falls short of accurately modeling the brain's capacity for 3D spatial imaging. 
Effective description of 3D objects necessitates the consideration of their shape and structural attributes, in addition to semantic properties.
For example, two cars may appear identical when viewed from the front, yet present substantial differences in length when observed from the side. 
This highlights the critical need to capture the full range of spatial and structural features for an authentic representation of 3D objects. 
Accordingly, our work aims to advance the modeling of human 3D comprehension. It necessitates the development of an advanced fMRI feature extractor capable of discerning not only semantic elements but also spatial structures and other 3D-specific characteristics from fMRI signals. This approach aims at enabling a more complete and accurate reconstruction of 3D visual information.

Addressing the unique challenges of our Recon3DMind task, we confront a significant hurdle: the absence of a dataset that pairs fMRI data with 3D visuals. 
To fill this void, we introduce fMRI-Shape, a pioneering resource specifically designed for this purpose, featuring data from 14 participants and supporting various experimental settings.
Inspired by ZeroNVS~\cite{sargent2023zeronvs}, which suggests that 3D objects can be effectively represented through 360-degree view videos, we adopt a similar methodology for our fMRI data collection. In our approach, participants are presented with 360-degree view videos of stationary 3D objects from ShapeNet~\cite{chang2015shapenet}. These videos feature the camera orbiting the object, completing a full rotation, thereby providing subjects with a comprehensive view of the objects from all possible angles. This method ensures the capture of detailed and precise fMRI signals as subjects observe the objects, effectively encapsulating the full range of the object's spatial attributes. Through meticulous preprocessing, these recordings are converted into multi-frame fMRI signals, capturing a rich dataset that allows for a nuanced analysis. The complexities and specific details of the fMRI-Shape dataset will be further elaborated in Sec.~\ref{sec:3}.

With our specially curated fMRI-Shape dataset, we present an innovative and effective three-stage framework to extract both spatial structure and semantic features from multi-frame fMRI signals, named Mind-3D. 

In the first stage, we utilize a pre-trained encoder~\cite{qian2023semantic} on the NSD~\cite{allen2022massive} dataset to extract spatial features from the fMRI data. These features are then aggregated across multiple frames using a feature aggregation module. To ensure the biological relevance and effectiveness of these features, we align them with the visual-spatial features of the corresponding objects using contrastive learning loss. This alignment guarantees that the features we extract are not only accurate but also meaningfully connected to the brain's visual processing.

The second stage focuses on decoding the brain's visual activity features from the extracted fMRI data. We achieve this through a transformer-based diffusion model trained within the feature space, conditioned on the fMRI features. This conditioning allows us to progressively generate accurate visual representations, effectively translating complex brain activity into comprehensible visual data.
 
The final stage aims to reconstruct the 3D models as they are perceived by the human brain. To accomplish this, we train a latent-adapted Argus~\cite{qian2024pushing} model. This model uses the visual features generated in the second stage along with the initially extracted fMRI features to create 3D objects. This process mirrors the imaging function of the brain's visual cortex, offering a simulated insight into how the brain perceives and processes 3D objects.

To evaluate the efficacy of our model, we employ various semantic and structural metrics. These metrics offer a comprehensive assessment of the model's ability to generate structurally accurate 3D representations. Additionally, we validate the relevance of the features extracted by our model by correlating them with specific brain regions. This ensures that our model not only produces accurate representations but also does so in a manner consistent with the brain's processing of visual information.

Our contributions can be summarized as follows:
\begin{itemize}[leftmargin=*,itemsep=0pt,topsep=0pt,parsep=0pt]
\item For the first time, we propose and demonstrate the feasibility of the novel task, \textbf{Recon3DMind}, which holds significance for both the computer vision and cognitive neuroscience communities.
\item We present \textbf{fMRI-Shape}, the first dataset of fMRI-3D shape pairs, across various settings, to foster further research in this domain. This dataset will be accessible on our project page.
\item We introduce an innovative, versatile brain decoding framework \textbf{MinD-3D}, composed of three distinct modules: a neuro-fusion encoder, a feature-bridging diffusion model, and a latent-adapted decoder.
\end{itemize}

\section{Related work}
\label{sec:Related_work}

\subsection{FMRI Decoding Methods}
Current fMRI decoding methods primarily focus on reconstructing the vision perception in a 2D format, such as the images or videos perceived by humans. This task is complex, as it requires accurately extracting relevant features from fMRI signals and reconstructing a precise 2D representation. Deep learning methods, known for their impressive capabilities, are well-suited to address this challenge. Initial successes in this area have been demonstrated by earlier methods~\cite{horikawa2017generic,wen2018neural,shen2019deep}. Subsequent studies~\cite{shen2019end, du2019brain} have shown that generative models are particularly effective for these tasks, leading to the employment of various diffusion models~\cite{chen2023seeing, chen2023cinematic, scotti2023reconstructing, sun2024contrast} as decoders to reconstruct visual scenes, achieving remarkable results. However, these studies have been limited to 2D visual representations and the related vision ROIs. In this paper, we aim to extend the scope of fMRI decoding to 3D representations, involving more vision ROIs. Our goal is to directly reconstruct 3D objects from fMRI signals. To accomplish this, we propose a new framework that employs a transformer-based feature encoder for extraction and aggregation. This framework translates neural space data into visual space and utilizes a powerful 3D decoder to reconstruct the 3D object, leveraging features from the visual space.

\subsection{Diffusion Models}
Diffusion models~\cite{ho2020denoising,song2020denoising} are exceptional generative tools for both pixel and feature generation. As a variant, the latent diffusion model~\cite{rombach2021highresolution}, equipped with an autoencoder, compresses images into lower-dimensional latent features, thereby generating a compressed version of the data rather than directly generating the data itself. Dit~\cite{peebles2023scalable} replaces the backbone of diffusion models with transformers, which will improve the performance and scalability of these models. This approach, operating in the latent space, significantly reduces computational requirements and enables the generation of higher-quality images with enhanced details in the latent space. In this paper, we aim to leverage the potent feature-generation capabilities of diffusion models to generate visual features based on fMRI features. To achieve this, we have adapted a transformer-based diffusion model, focusing solely on its latent component. The conditional information driving the model is derived from the fMRI features.

\subsection{3D Generation}
3D generation can be accomplished through various methods~\cite{qian2024pushing,zhuo2024vividdreamerinvariantscoredistillation,tang2024lgm,liu2024mirrorgaussian}. Some methods~\cite{tang2023dreamgaussian, tang2024lgm} employ 3D Gaussian splatting~\cite{kerbl3Dgaussians} for this purpose.      
Other studies~\cite{wang2023imagedream, ye2024dreamreward} utilize diffusion models to generate multi-view representations of objects, subsequently constructing 3D models. Additionally, traditional and direct approaches leverage autoregressive methods~\cite{cheng2022autoregressive, ibing2023octree, qian2024pushing} for 3D object generation.
In our study, we aim to generate 3D objects from fMRI data, using Argus~\cite{qian2024pushing}, a robust 3D generative model, as our decoder, which we enhance with several transformer layers. This approach integrates fMRI and visual features produced by the preceding module. The visual features act as conditional embeddings for Argus, while the fMRI features are fused within the adapter layers. This synergistic integration seeks to augment the model's capability to accurately reconstruct 3D objects from complex brain activities.

\begin{figure}[tb]
    \centering
    \includegraphics[width=0.55\linewidth]{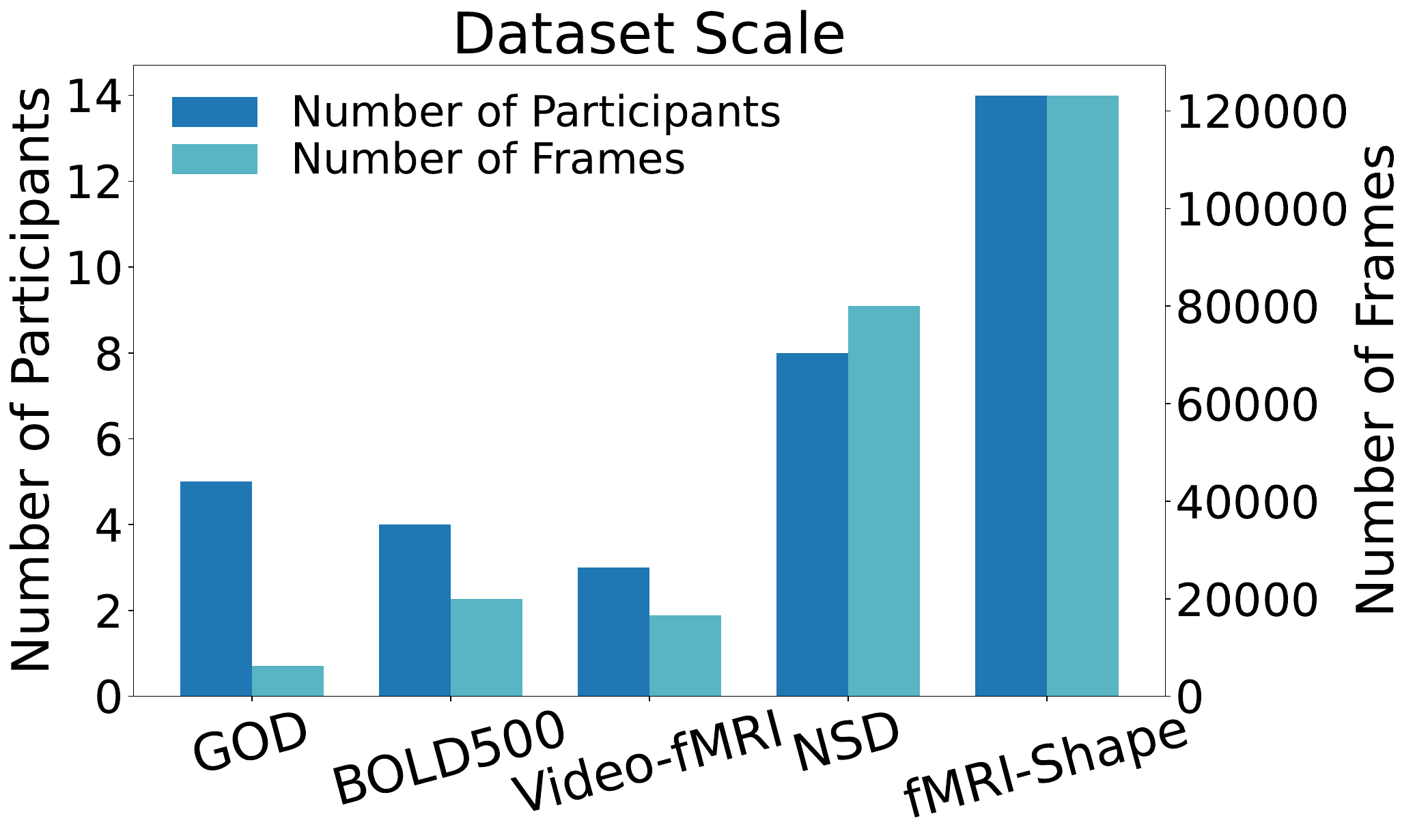}
    \caption{\textbf{Comparing fMRI-Shape with other 2D fMRI datasets.} As the first 3D fMRI dataset, fMRI-Shape features a larger number of participants and frames, providing ample support for experiments in our proposed novel task and further research.
    \label{fig:dataset_scale}
    }
\end{figure}

\begin{table}[tb]
    \centering
    \caption{
        \textbf{Details of fMRI-Shape Dataset.} It ensures a balanced representation of male and female participants. Specifically, the Across-Person Set (AP Set) includes fMRI data sampled from different participants for Core test data. The Across-Person \& Across-Class Set (APAC Set) involves different participants viewing new 3D objects. These two subsets facilitate evaluations for model generalization.
        \label{tab:detail_about_dataset}
    }
    \setlength{\tabcolsep}{1.5mm}{
    \begin{tabular}{lccccc}
    \toprule
      &  Participant & Males/Females & Category  & Objects & Frames \\
    \midrule
    \midrule
    Core Set   & 8    &  4/4   & 13   &  1404 & 14040 \\
    AP Set & 2    &  1/1   & 13   &  104  & 1040 \\
    APAC Set & 4    &  2/2   & 55   &  220  & 2200 \\
    \midrule
    Total & 14 & 7/7 & 55 & 1624 & 123200\\
    \bottomrule
    \end{tabular}
    }
\end{table}

\section{Experimental Designs and Curated Dataset}
\label{sec:3}
In this section, we will introduce the detailed procedure for collecting the proposed dataset, fMRI-Shape. 
The scale of fMRI-Shape compared to other datasets (NSD~\cite{allen2022massive}, BOLD5000~\cite{chang2019bold5000}, GOD~\cite{horikawa2017generic}, Video-fMRI~\cite{wen2018neural}) is illustrated in Fig.~\ref{fig:dataset_scale}, while the specific details of fMRI-Shape are provided in Tab.~\ref{tab:detail_about_dataset}.
This dataset are released to the community to push the research on the topic of Recon3DMind.

\begin{figure}[t]
    \centering
    \includegraphics[width=\columnwidth]{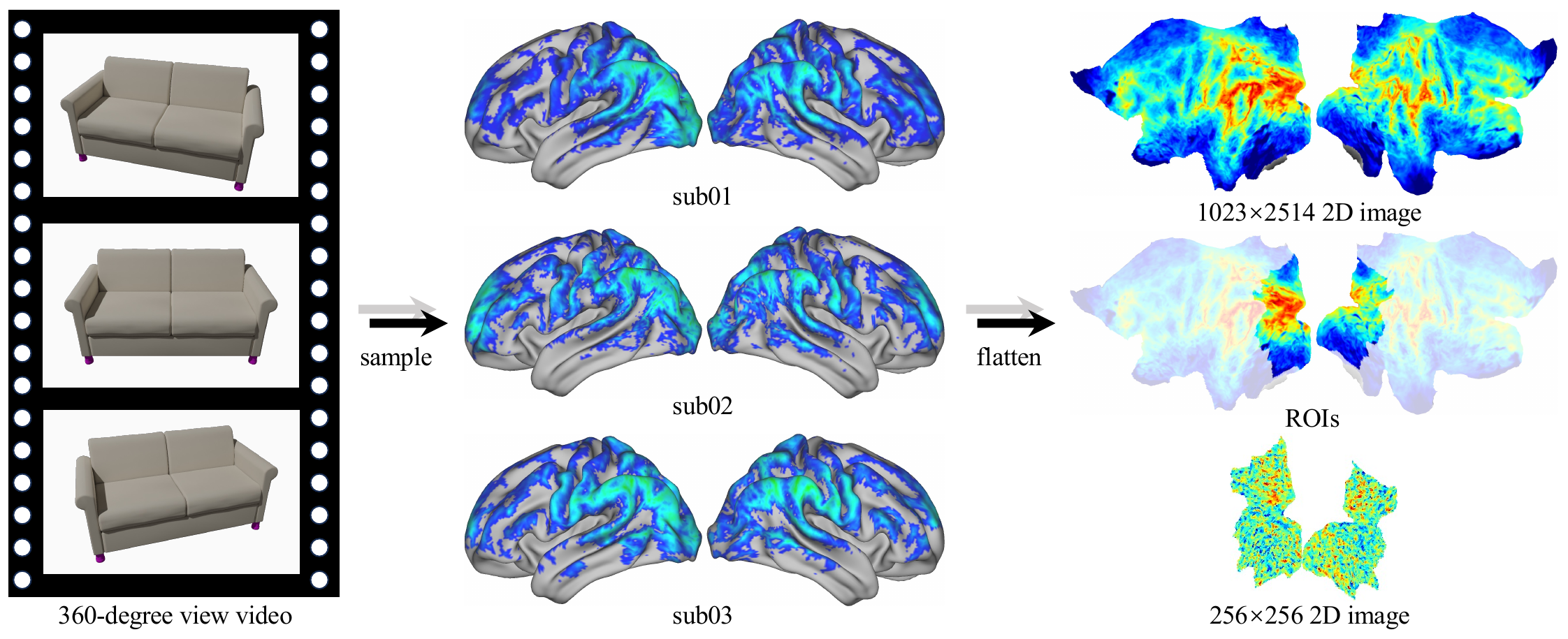}
\caption{
\textbf{Overview of the fMRI-Shape Acquisition Process.} Initially, we render each object into an 8-second long video, showcasing a 360-degree view. Subsequent fMRI signal capture is performed in video format, followed by data processing with fMRIPrep to convert signals from 32k\_fs\_LR surface space into 2D images of dimensions 1023 $\times$ 2514. \textbf{Individual differences} observed in the dataset, as highlighted in the middle part, underscore the challenges in generalizing these findings. On the rights, regions of interest (ROIs) are transformed into 256 $\times$ 256 image.
}
\label{fig:dataset_example}
\end{figure}

\subsection{Subjects and Experiments}
FMRI-Shape encompasses data from 14 participants unfamiliar with the objectives of our study. All participants possessed normal or corrected-to-normal visual acuity. Informed written consent was obtained from each participant, and the experimental protocol received approval from the ethical review board. 

To ensure the diversity of the dataset, the 3D objects used in this study were derived from the ShapeNetCore~\cite{chang2015shapenet}, which includes 55 classes of objects. We utilized the rendering technique from Zero123~\cite{liu2023zero1to3} to generate an 8-second video for each object using Blender. These videos illustrate the gradual 360-degree rotation of 3D objects at a 60-degree pitch angle, as exemplified in Fig.~\ref{fig:dataset_example}. 

\noindent \textbf{Core} The core set of fMRI-Shape includes data from 8 participants: 4 males and 4 females, aged 21 to 29 (Participants No. 1-8). We selected a total of 1404 objects across 13 common categories frequently employed in various 3D reconstruction literature~\cite{chen2019learning, sun2020pointgrow,ibing20213d} within ShapeNetCore. For each category, 100 objects were allocated for training and 8 for testing, totaling 108 objects per category. The experiment lasted approximately 4.5 hours per participant, divided into 32 sessions. In each session, participants viewed 48 videos in a randomized order, interspersed with 2.4 second rest periods, covering one object from each category, resulting in a session duration of around 9 minutes. Sessions alternated between training and test data, with a 1-hour break after every 8 sessions to prevent fatigue and maintain data quality. Notably, the testing objects were viewed twice by each participant.

\noindent \textbf{Across-Person Set (AP Set)} The AP Set, designed for Out-Of-Distribution (OOD) testing, includes fMRI data from 2 participants (1 male aged 24 and 1 female aged 26, Participants No. 9 and 10), sampled while they were viewing the test objects from the Core set.

\noindent \textbf{Across-Person \& Across-Class Set (APAC Set)}
The APAC Set represents a more challenging component than the AP Set. It includes data from 4 participants: 2 males and 2 females, aged 22 to 26 (Participants No. 11-14). We randomly selected 4 objects from each of the 55 categories in ShapeNetCore, distinct from those in the Core, for a total of 220 objects. Each participant was exposed to these 220 objects across 6 sessions.

As shown in the middle part of Fig.~\ref{fig:dataset_example}, we can observe individual differences, which will pose significant challenges for generalization. Thus, the AP and APAC sets are crucial for OOD testing and will serve as important benchmarks to assess the generalization capabilities of the 3D decoding model.

\subsection{Data Acquisition and Preprocessing}
T1 and fMRI data were acquired in a 3T scanner and a 32-channel RF head coil. T1-weighted data were scanned using MPRAGE sequence (0.8-mm isotropic resolution, TR=2500ms, TE=2.22ms, flip angle $8^{\circ}$). Functional data were scanned using gradient-echo EPI at 2-mm isotropic resolution with whole-brain coverage (TR=800ms, TE=37ms, flip angle $52^{\circ}$, multi-band acceleration factor 8). The sampling frequency of the 3T scanner is 1.25Hz, so each video segment corresponds to a total of 10 frames of task-state fMRI signals.

Stimuli were presented using an LCD screen ($8^{\circ}\times 8^{\circ}$) positioned at the head of the scanner bed. Participants viewed the monitor via a mirror mounted on the RF coil and fixated a red central dot (0.4° × 0.4°). 

Preprocessing was performed using fMRIPrep~\cite{fmriprep1,fmriprep2}. Following~\cite{qian2023fmri}, the preprocessed functional data in 32k\_fs\_LR surface space were converted into 2D images and utilized for further analysis. Given the delay of the BOLD signal by 6 seconds, we applied z-scoring to the data points across every vertex within each run, incorporating a 6-second lag. These normalized values were then projected onto 1023 × 2514 pixel 2D images using pycortex. For analysis, Regions of Interest (ROIs) were selected from the Human Connectome Project Multi-Modal Parcellation (HCP-MMP) atlas in the 32k\_fs\_LR space. These ROIs included areas such as ``V1, V2, V3, V3A, V3B, V3CD, V4, LO1, LO2, LO3, PIT, V4t, V6, V6A, V7, V8, PH, FFC, IP0, MT, MST, FST, VVC, VMV1, VMV2, VMV3, PHA1, PHA2, PHA3''. Subsequently, the ROIs were converted into a 256 × 256 image, as illustrated in the right part of Fig.~\ref{fig:dataset_example}.

\section{Problem Setup}

Recon3DMind addresses a task of paramount importance in cognitive neuroscience. Developing computational models that accurately interpret and reconstruct the human brain's 3D comprehension can significantly advance the field of computer vision. This endeavor not only bridges a gap between cognitive neuroscience and computer vision but also promises to propel the latter field forward in unprecedented ways. In this paper, we delve into the specifics of fMRI-based 3D reconstruction, providing detailed definitions and formulas underpinning our methodology.

Consider the acquisition of a multi-frame fMRI signal denoted as $\{F\}$, where $|F|=10$. These signals correspond to both a 3D object mesh, $\Psi$, and a video, \(\{V\}\), with $|V|=192$ frames, which the subject views. The task necessitates an efficient encoder, $E$, capable of extracting both spatial structural and semantic features from the fMRI signal. It's crucial to note that while a single frame of the fMRI signal suffices for semantic information extraction in 2D image reconstruction, the reconstruction of 3D structures demands additional spatial structural features. Therefore, we input multiple frames of fMRI signals with the aim of extracting these comprehensive spatial structural features from the spatio-temporal data. This process is mathematically represented as: $ f = E(F) $ Following the feature extraction, we employ a powerful decoder to reconstruct the original 3D mesh $\Psi$ based on the extracted feature $f$:  $\Psi = D(f) $
Consequently, our model can be succinctly defined as $ M = \{E, D\}$, thereby operationalizing the transformation: $\Psi = M(F) $.

In executing this model, it's critical to fully utilize both the fMRI signal $\{F\}$ and the corresponding video $\{V\}$ to train our model $M$. To achieve this, we propose a three-stage innovative and efficient model. Each stage is meticulously designed to capture different aspects of the fMRI data and the associated visual stimuli, ensuring a comprehensive and accurate 3D reconstruction from the complex neural signals. This process not only tests the limits of current computer vision techniques but also offers valuable insights into the human brain's processing of 3D spatial information.


\section{Method}

\begin{figure}[t]
    \centering
    \includegraphics[width=\columnwidth]{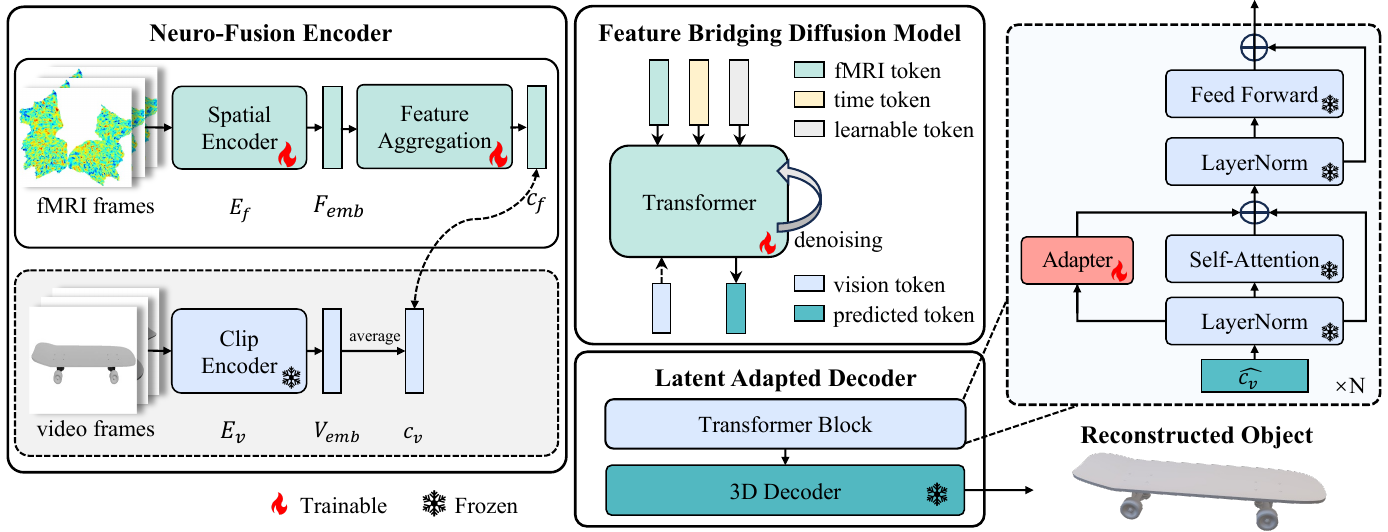}
\caption{
\textbf{Overview of the MinD-3D Framework.} Our approach combines a Neuro-Fusion Encoder for extracting features from fMRI frames, a Feature Bridge Diffusion Model for generating visual features from these fMRI signals, and a Latent Adapted Decoder based on the Argus 3D shape generator for reconstructing 3D objects. This integrated system effectively aligns and translates brain signals into accurate 3D visual representations. Note that the CLIP encoder is only for training the model, while not used for inference.
}
\label{fig:framework}
\end{figure}

\subsection{Overview}

As depicted in Fig.~\ref{fig:framework}, our MinD-3D model comprises three main components: \textbf{1) Neuro-Fusion Encoder:} This part is responsible for extracting both semantic and structural features from the fMRI frames. \textbf{2) Feature Bridge Diffusion Model:} This module generates visual features based on the features extracted by the Encoder. \textbf{3) Latent Adapted Decoder:} Finally, this component recovers 3D objects, using both the fMRI features and the generated visual features.

\subsection{Neuro-Fusion Encoder}
Drawing from the potent Masked Image Model (MIM)~\cite{he2022masked, gao2023coarse, qian2023semantic}, we adopt an auto-encoder structure to extract meaningful fMRI features, trained on the UKB dataset~\cite{miller2016multimodal}. 
Our encoder, termed the Neuro-Fusion Encoder (NFE), employs the encoder of LEA~\cite{qian2023semantic} as $E_{f}$, along with a Feature Aggregation module ($\mathcal{FA}$), to transform fMRI frames into embeddings. Inspired by contrastive learning~\cite{radford2021learning,cherti2023reproducible,zhuo2023whitenedcse}, we then align these embeddings with the vision space through CLIP~\cite{radford2021learning}.

Initially, we process each fMRI frame in parallel to obtain the spatial fMRI embeddings:
$$\mathbf{F}_{emb} = E_{f}(\mathbf{F})$$
Then, we use the Feature Aggregation module to aggregate the fMRI embeddings into fMRI latent feature $\mathbf{c}_f$:
$$\mathbf{c}_f = \mathcal{FA}(\mathbf{F}_{emb})$$
To align the fMRI latent feature with the vision space, we employ the ViT-B/32 CLIP vision encoder $E_{v}$ and here to encode video frames into vision embeddings. During training, we randomly select n frames as $\{\mathbf{V}\}$ and compute the average vision embeddings $V_{emb}$ as the visual latent feature $\mathbf{c}_v$:
$$\mathbf{c}_v = \frac{\sum^{n}_{i=1} E_{v}(\mathbf{V_{i}})}{n}$$
where n is the number of selected video frames. Then we utilize the CLIP loss to ensure the alignment between fMRI features $\mathbf{c}_f $ and visual features $\mathbf{c}_v$:
$$\mathcal{L}_c = \mathcal{L}_{clip}(\mathbf{c}_f, \mathbf{c}_v)$$
During training, we optimize the neuro-fusion encoder initialized with pre-trained weights, while the CLIP vision encoder remains frozen. It's worth noting that the CLIP vision encoder $E_{v}$ and images $\{\mathbf{V}\}$ are just used for training, we will drop it during inference. By adopting this approach, we will get the fMRI features $\mathbf{c}_f$ and embedding $\mathbf{F}_{emb}$ aligned with the visual space. The fMRI features $\mathbf{c}_f$ will be used as the conditional information for the feature bridge diffusion model.

\subsection{Feature Bridge Diffusion Model}
This section describes the Neural-Visual Synthesis process via a transformer-based diffusion model, herein referred to as the Feature Bridge Diffusion Model (FBDM), inspired by the hierarchical design proposed in~\cite{ramesh2022hierarchical}. The FBDM's primary role is to bridge fMRI latent features $\mathbf{c}_f$ with their visual counterparts $\hat{\mathbf{c}_v}$, effectively transforming neurological signals into visual representations.
In the forward diffusion process, we treat the visual latent feature $\mathbf{c}_v$ as $\mathbf{x}_0$ and introduce Gaussian noise into it across 100 timesteps. This approach modifies the standard 1000-timestep model to align with the settings in Mind-eye~\cite{scotti2023reconstructing}:
$$ \mathbf{x}_t = \sqrt{\alpha_t} \mathbf{x}_{t-1} + \sqrt{1 - \alpha_t} \mathbf{\epsilon}, \quad \mathbf{\epsilon} \sim \mathcal{N}(0, \mathbf{I}) $$
During the reverse diffusion process, FBDM is tasked with predicting and reversing the introduced noise to reconstruct the latent visual feature. This is achieved by estimating the noise $\hat{\mathbf{\epsilon}}_t$ at each timestep, conditioned on the fMRI latent feature $\mathbf{c}_f$:
$$ \hat{\mathbf{\epsilon}}_t = FBDM(\mathbf{x}_t, t, \mathbf{c}_f) $$
Differing from U-Net based models, our approach utilizes a transformer to leverage the self-attention mechanism for integrating the conditional information $\mathbf{c}_f$. This is done by embedding the timestep $t$, the noise-perturbed feature $\mathbf{x}_t$, and the conditional fMRI feature $\mathbf{c}_f$ into a unified representation, further augmented with a learnable token to capture the target vision features accurately.
The objective of FBDM training is to minimize the discrepancy between the actual noise and its prediction:
$$ \mathcal{L}_{FBDM} = \mathbb{E}_{\mathbf{x}, \mathbf{\epsilon}, t} \left[ \| \mathbf{\epsilon} - \hat{\mathbf{\epsilon}}_t \|^2 \right] $$
Upon training completion, the model discards the initial visual latent feature $\mathbf{c}_v$ and enters a generative phase. It begins with pure noise and progressively refines this through the reverse diffusion process to generate the predicted visual latent feature $\hat{\mathbf{c}_v}$. This process effectively translates the aligned fMRI latent features $\mathbf{c}_f$ into visual representations, serving as essential conditional information for the subsequent latent-adapted decoding phase.

\begin{figure}[t]
    \centering
    \includegraphics[width=\columnwidth]{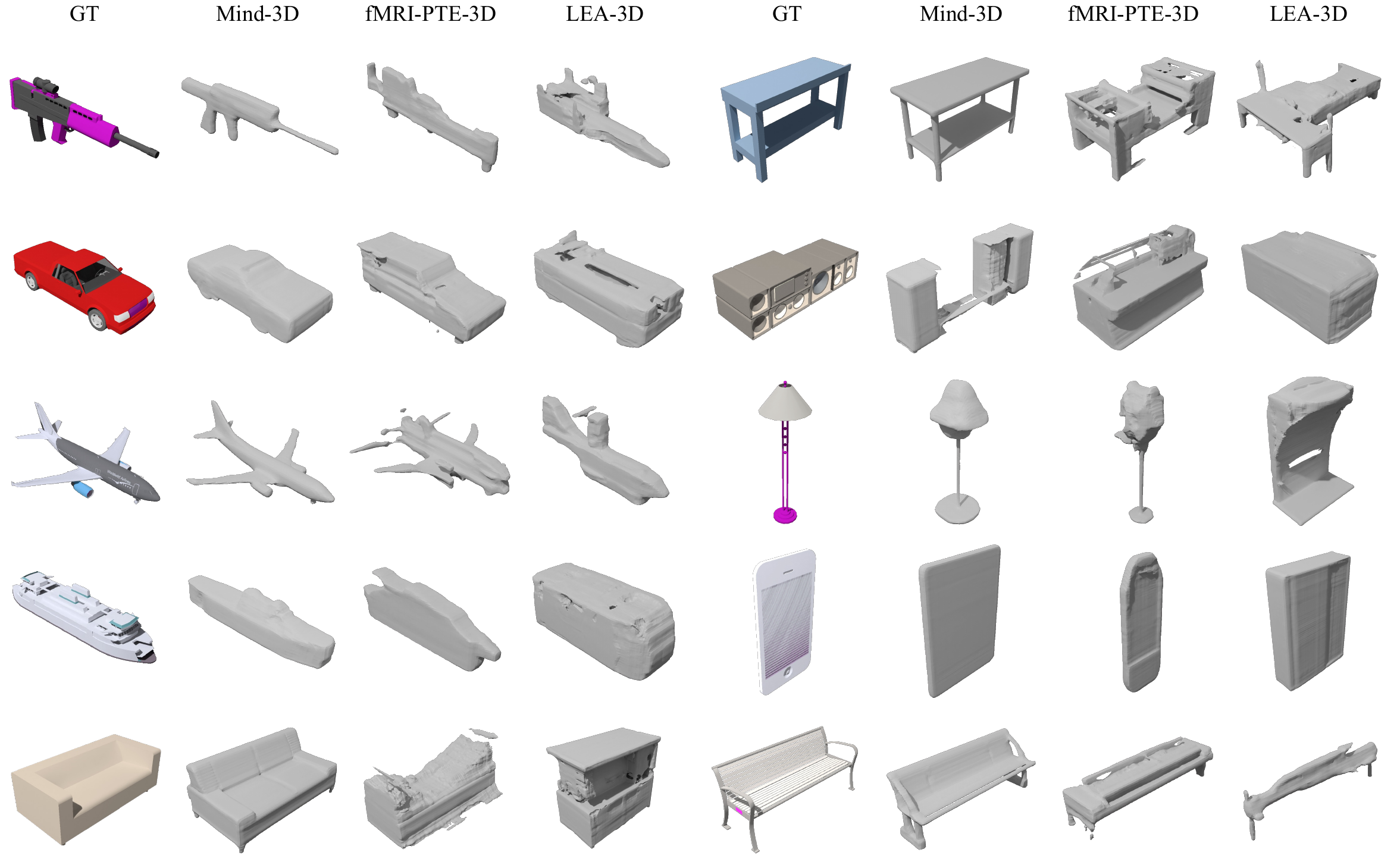}
\caption{
\label{fig:visualize}
The qualitative results generated by LEA-3D, fMRI-PTE-3D, and our method are presented. GT indicates the ground-truth 3D objects. All the objects have been rendered into a 2D format.
}
\end{figure}

\subsection{Latent Adapted Decoder}
The Latent Adapted Decoder (LAD) leverages Argus~\cite{qian2024pushing}, a state-of-the-art 3D mesh generation method, as its backbone. This component is tasked with generating the Vector Quantized (VQ) latent representation of target 3D objects from conditional information processed through a generative auto-regressive transformer. The resulting 3D objects are then reconstructed using the VQ decoder. 

To tailor Argus for processing fMRI input and to retain its generative efficiency, we employ adapter tuning during training. This involves integrating the adapter layers within the transformer block. Following classic methods~\cite{houlsby2019parameter,yang2023aim}, we apply two feed-forward linear layers to effectively adapt the self-attention outputs.
The adapter layer is systematically inserted at every $T$ transformer blocks, ensuring that the model remains adaptable and efficient. The decoder formulates the VQ latent vector for the target object as follows:
$$
p(\mathbf{z}) = \prod_{i=1}^{m} p\left(\mathbf{z}_i \mid \hat{\mathbf{c}_v}, \mathbf{z}_{<i}\right)
$$
Subsequently, the 3D VQ-decoder decodes this latent vector $\mathbf{z}$ into the final 3D mesh representation $\Psi$:
$$
\Psi = D(\mathbf{z})
$$
The decoder aims to minimize the negative log-likelihood loss during optimization:
$$
\mathcal{L}_{nll} = \mathbb{E}_{\mathbf{x} \sim p(\mathbf{x})}\left[-\log p(\mathbf{z})\right]
$$
For training, the decoder is fine-tuned from pre-trained weights, with the trained neuro-fusion encoder and the feature bridge diffusion model. This strategy preserves Argus's generative capabilities while adapting to new inputs and minimizing the computational overhead associated with training.

\begin{table}[tb]
\caption{
\textbf{Performance Comparison on fMRI-Shape.} We report the average metrics for each subject trained and tested on their own data in the core part. LEA-3D and fMRI-PTE-3D represent variants of LEA and fMRI-PTE, respectively. The following three baselines are ablation studies for MinD-3D.
\label{tab:metric_table}
}
\centering{
    \setlength{\tabcolsep}{1.5mm}{
    \begin{tabular}{l|ccc|cccc}
    \toprule
    \multicolumn{1}{c|}{\multirow{2}{*}{\textsc{Methods}}}   & \multicolumn{3}{c|}{Semantic-Level} & \multicolumn{4}{c}{Structure-Level}\\
        & 2-way$\uparrow$  &  10-way$\uparrow$    & LPIPS$\downarrow$  & SSIM$\uparrow$ & FPD$\downarrow$ & CD$\downarrow$  & EMD$\downarrow$     \\
    \midrule
    \midrule
    LEA-3D\cite{qian2023semantic} &    0.787      &   0.371   &      0.527   &  0.562   &  4.229     &  2.291  & 5.347    \\
    fMRI-PTE-3D\cite{qian2023fmri}&    0.815      &   0.392   &     0.433   &  0.694   &  3.571     &  1.992  & 4.621    \\
    \midrule
    w/o Both          &      0.789      &  0.367                  &  0.479   &  0.616   &  3.694     &  2.205  & 5.073    \\
    w/o Diffusion     &      0.801      &  0.385                  &  0.423   &  0.669   &  3.526     &  2.071  & 4.625    \\
    w/o Contrastive   &      0.823      &  0.419                &  0.319   &  0.701   &  3.315     &  1.826  & 4.027    \\
    MinD-3D (full)    & \textbf{0.839}  & \textbf{0.432}  &  \textbf{0.230}  & \textbf{0.734} & \textbf{3.157}  & \textbf{1.742} & \textbf{3.833}  \\ 
    \bottomrule
    \end{tabular}}
    }
\end{table}

\section{Experiments}

\subsection{Metrics} 
To more effectively evaluate the performance of our models on the new task, we utilize metrics in two aspects: semantic-level and structure-level.

\noindent \textbf{Semantic Level.} To evaluate the semantic quality of our model, we adopt standard metrics used in previous 2D fMRI literature~\cite{chen2023seeing,chen2023cinematic,ozcelik2022reconstruction,mai2023unibrain,sun2024contrast}, namely, N-way top-K accuracy. We calculate 2-way-top-1 and 10-way-top-1 accuracy as shown in Tab.~\ref{tab:metric_table}. Additionally, we calculate the Learned Perceptual Image Patch Similarity (LPIPS)~\cite{zhang2018perceptual} to assess the perceptual quality of semantics. In practice, these two metrics are calculated by comparing the reconstructed images with the ground truth (GT) images. In this process, we render both the reconstructed and GT objects into images separately at every 60 degrees, calculate the metrics for each frame, and then derive the average value as the final metric numbers.

\noindent \textbf{Structure Level.}  
In addition to the semantic quality evaluation, it is crucial to assess the performance of our model in capturing geometrical structures. We utilize common metrics employed in 3D reconstruction~\cite{liu2022towards,xu2019disn,qian2024pushing}: Fréchet Point Cloud Distance (FPD, adjusted by a factor of $\times 10^{-1}$), Chamfer Distance (CD, scaled by $\times10^2$), and Earth Mover’s Distance (EMD, also scaled by $\times10^2$). We sample point clouds from the GT and reconstructed meshes to calculate these three metrics. Furthermore, we compute the frame-averaged Structural Similarity Index (SSIM) using the same rendering method to assess similarity from each viewpoint.

\subsection{Implementation Details}
As detailed in Sec.~\ref{sec:3}, each data pair in the fMRI-Shape dataset is composed of 10 fMRI frames and 192 corresponding images. To maximize dataset utilization and introduce data augmentation, we randomly select 6 fMRI frames for training. These frames, originally 1023 × 2514 2D fMRI images, are resized to 256 × 256. 
For the image data, 8 frames are randomly selected for training, with each image resized to 224 × 224. During inference, only the middle 6 fMRI frames are used, without incorporating additional images.
Regarding the architecture, we set $T=4$ to incorporate adapter layers.

The model training proceeds in two stages. Initially, we focus on optimizing the neuro-fusion encoder and the feature bridge diffusion model. The neuro-fusion encoder is initialized with pre-trained weights, while the feature bridge diffusion model undergoes training from scratch, requiring approximately 2.5 days on a single A100 GPU. In the second stage, we optimize all parameters, including those of the adapters, and freeze the parameters in Argus. This stage is completed in about one day on a single A100 GPU. Details on the hyperparameters of the model are provided in the supplementary material.
\subsection{Experimental Results}
As the pioneering effort in modeling 3D imaging within the human brain, which involves various regions of the brain, direct comparisons with existing models are not feasible. To mitigate the training costs, we have adapted LEA and fMRI-PTE models, which were trained on the same vision ROIs, utilizing the identical 3D decoder as employed in MinD-3D. These adaptations serve as the baselines in this study, facilitating a more contextual and fair comparison within the novel domain we are exploring. Tab.~\ref{tab:metric_table} presents the averaged metrics at both the Structural and Semantic Levels for all subjects. MinD-3D demonstrates superior performance across all metrics: achieving approximately 83.9\% accuracy in 2-way-top-1, 43.2\% accuracy in 10-way-top-1, and scores of 0.734, 0.230, 3.157, 1.742, and 3.833 in other metrics. These results indicate that our model not only generates objects with high semantic accuracy but also excels in structural similarity, outperforming baseline methods. Tab.~\ref{tab:metric_table} also proves the effectiveness of the diffusion model and the contrastive learning module, both of which play important roles in our MinD-3D.

Qualitative results from LEA-3D, fMRI-PTE-3D, and our method are showcased in Fig.~\ref{fig:visualize}. MinD-3D consistently produces 3D objects that are structurally very similar to their real counterparts, while maintaining semantic integrity in most instances. A notable example is the follower vase case, where despite the complexity of its semantics and structure, our model effectively captures and reproduces key features. This underscores the robustness of our model in handling this novel task and its capability to produce faithful realizations.

\subsection{Further Analysis}

To effectively utilize a subset of the fMRI-Shape dataset and assess the generalization capabilities of our proposed MinD-3D model, we conducted two Out-Of-Distribution (OOD) experiments under challenging settings:
\textbf{1) Across-Person Testing (APT):} In APT, we evaluate our model, which was trained only on Subject 1, using the data from Subject 9. We compare the results with the baselines and report the metrics in Tab.~\ref{tab:AP_APAC}.
\textbf{2) Across-Person \& Across-Class Testing (APACT):} In APACT, we similarly evaluate our model, trained solely on Subject 1, with the data from Subject 11. We also compare with the baselines and report the metrics in Tab.~\ref{tab:AP_APAC}.

Although the performance in these OOD scenarios does not match the In-Distribution (ID) results, which is to be expected for such an extremely challenging task and given the substantial individual differences and domain gaps, our method still outperforms the baselines. This achievement not only underscores the robustness of MinD-3D but also establishes a new benchmark for the community to build upon.

\section{Conclusion}

\begin{table}[tb]
  \caption{Quantitative results of APT and APACT. We use the model trained on Subject 1 and compare metrics for Subjects 9 and 11 separately.}
  \label{tab:AP_APAC}
  \centering
    \setlength{\tabcolsep}{1.5mm}{
        \begin{tabular}{l|ccc|ccc}
        \toprule
         \multicolumn{1}{c|}{\multirow{2}{*}{\textsc{Methods}}} & \multicolumn{3}{c|}{APT} & \multicolumn{3}{c}{APACT} \\
         & FPD$\downarrow$ & CD$\downarrow$  & EMD$\downarrow$  & FPD$\downarrow$ & CD$\downarrow$  & EMD$\downarrow$  \\
        \midrule
        \midrule
            LEA-3D          & 5.362          & 3.627              & 6.174  & 6.958          & 4.944              & 8.107  \\
            fMRI-PTE-3D     & 4.501          & 2.956              & 5.772  & 6.261          & 4.570              & 7.843  \\
            MinD-3D  & \textbf{3.838} & \textbf{2.415}   & \textbf{5.117}  & \textbf{5.689} & \textbf{4.181}   & \textbf{7.194}   \\ 
        \bottomrule
        \end{tabular}
    }
\end{table}

In this paper, we introduce the innovative task of Recon3DMind and its accompanying large-scale dataset, fMRI-Shape, across various settings for the first time. Technologically, we develop a novel three-stage framework that involves more brain regions, including those associated with 3D vision in humans, specifically designed for this task. This approach sets a new benchmark in the field and proves the feasibility of the task. Initially, our model proficiently extracts features from fMRI frames. It then utilizes a diffusion module in the second stage to transition these fMRI features into the vision domain. The culmination of this process is the transformation of vision features into 3D models by an advanced 3D generation model in the final stage. Our comprehensive experimental results and analyses affirm the model's effectiveness in precisely extracting fMRI features and converting them into their corresponding 3D objects. This pioneering work not only carves out a new niche in neuroimaging and 3D reconstruction but also paves the way for future research aimed at a deeper understanding and visualization of neural representations in 3D vision.

\section*{Acknowledgements}
The computations in this research were performed using the CFFF platform of Fudan University.

\bibliographystyle{splncs04}
\bibliography{main}

\counterwithin{figure}{section}
\counterwithin{table}{section}

\newpage

\appendix

\section{FMRI-Shape Details}
\subsection{Visualization of fMRI Across Different Subjects}
\begin{figure}[hb]
    \centering
    \includegraphics[width=0.9\linewidth]{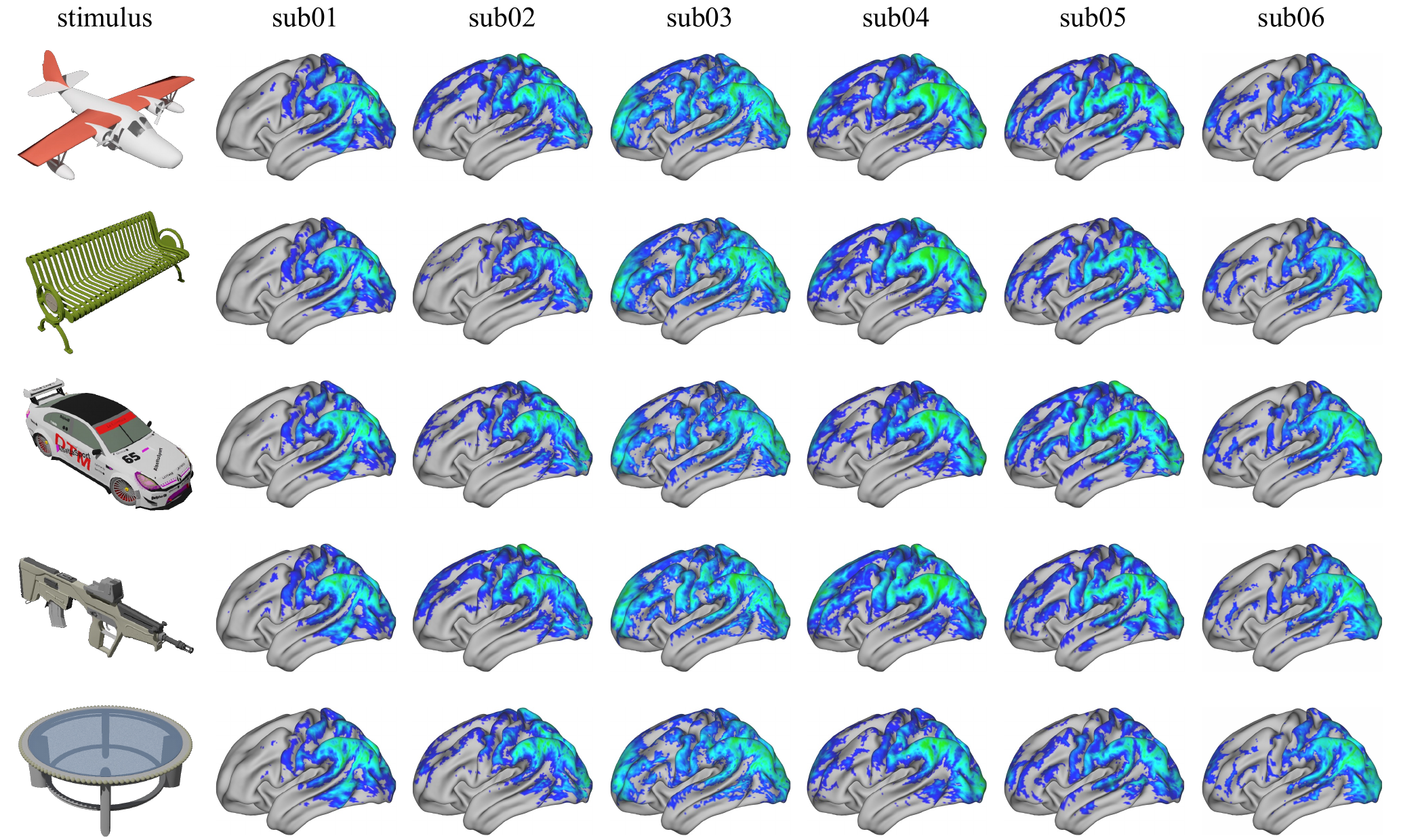}
\caption{
Individual differences in brain activation patterns within the fMRI-Shape.
\label{fig:fmri}
}
\end{figure}

To more effectively demonstrate brain activation patterns and showcase our fMRI-Shape dataset, we analyze and visualize the responses to five distinct objects across six different subjects in Fig.~\ref{fig:fmri}. (Note: Only voxels with activation levels exceeding the 50th percentile are displayed.)

Fig.~\ref{fig:fmri} also reveals significant individual differences in brain activation across subjects, more so than the variations in responses to different objects. This observation underscores the challenges inherent in the settings of AP and APAC.

\subsection{Visualization of the Core Component}
\begin{figure}[hb]
    \centering
    \includegraphics[width=0.95\linewidth]{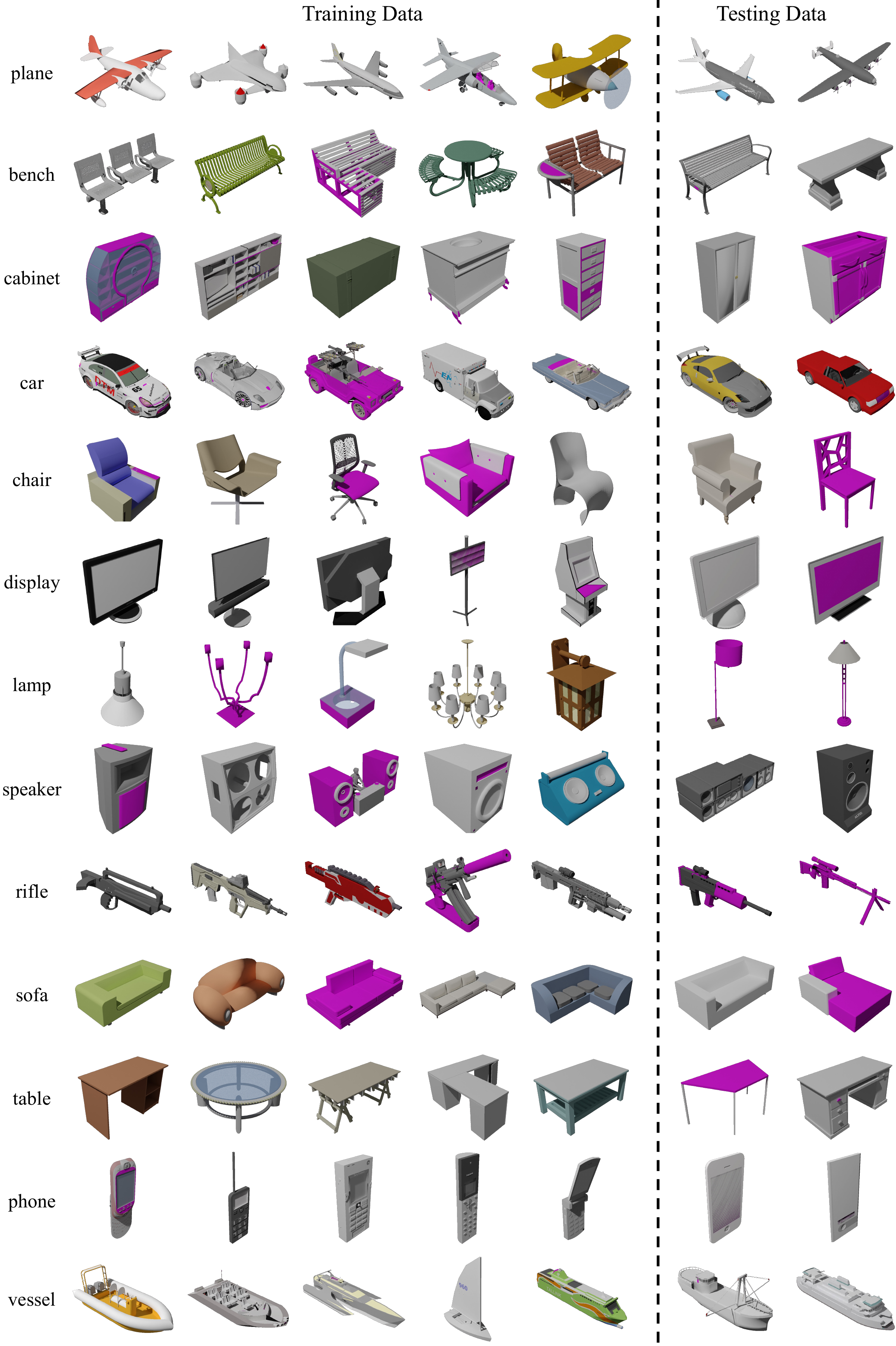}
\caption{
\textbf{Samples of Core component in fMRI-Shape.} We show five objects of training data and two objects of testing data per category.
\label{fig:Core}
}
\end{figure}

To illustrate the diversity of fMRI-Shape, Fig.~\ref{fig:Core} shows the core part, organized with five objects of training data and two objects of testing data each category.

\subsection{Visualization of the APAC Set}
\begin{figure}[t]
\centering
\includegraphics[width=0.95\linewidth]{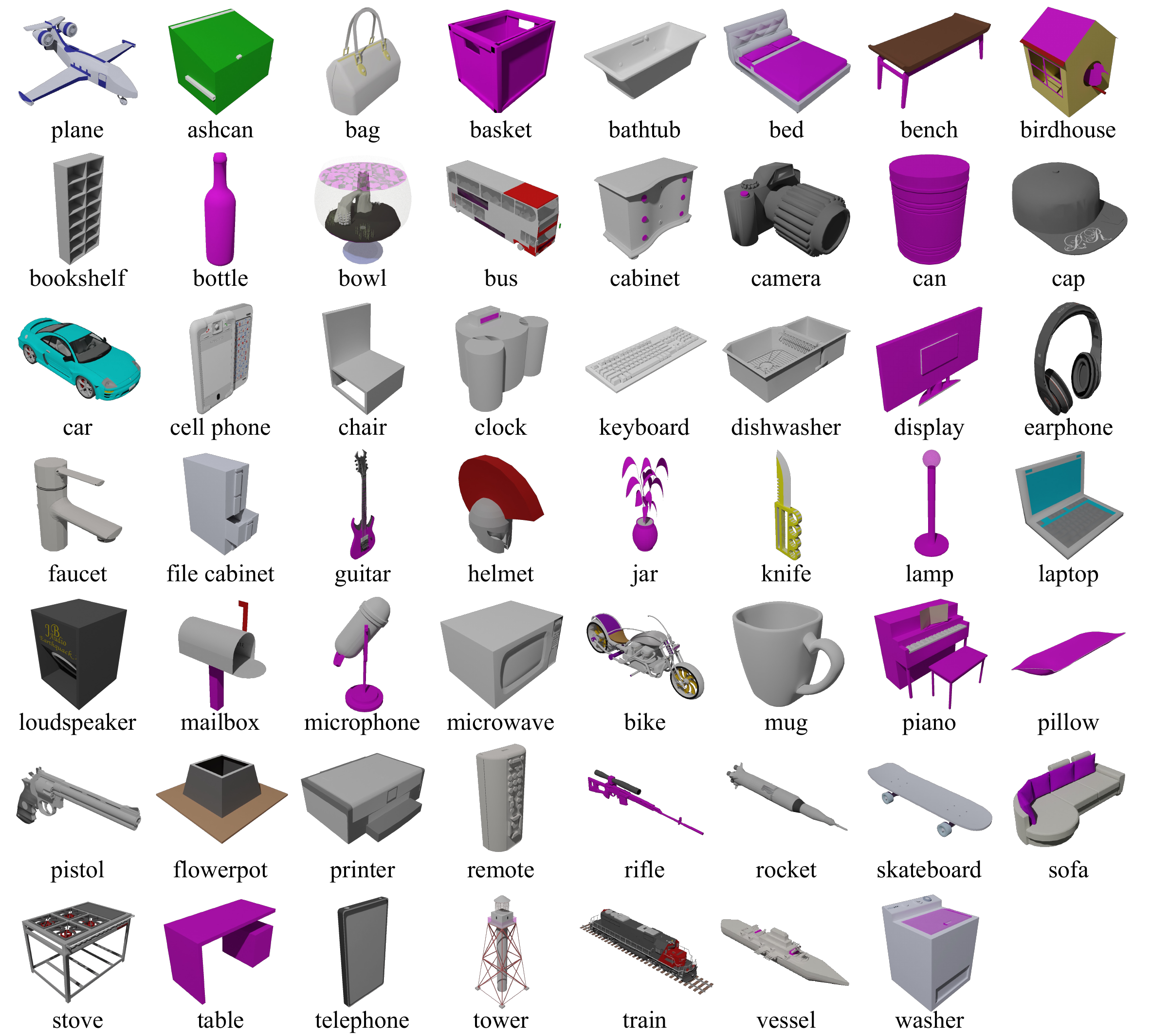}
\vskip -0.05in
\caption{
\textbf{Samples of APAC Set in fMRI-Shape.} One object is shown per category.
\label{fig:APAC}
}
\vskip -0.1in
\end{figure}

The APAC Set, which includes additional objects not existing in the Core parts, is showcased in Fig.~\ref{fig:APAC}, with one object displayed for each category.

\clearpage

\clearpage

\section{Analysis of Visualizations}

\subsection{Visualizations from Different Subjects}
\begin{figure}[tb]
\centering
\includegraphics[width=0.87\linewidth]{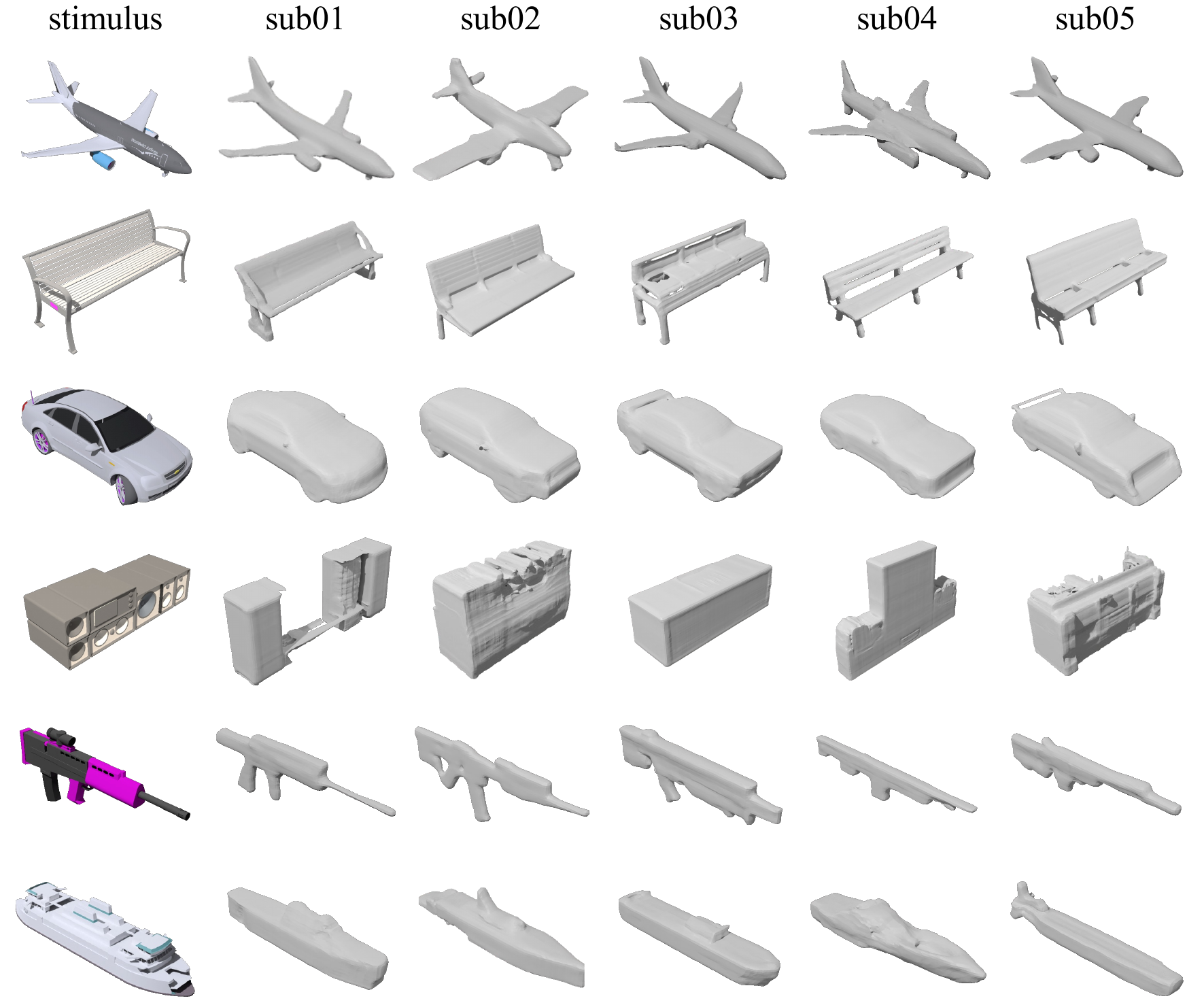}
\caption{
\textbf{Samples from different subjects.}
\label{fig:vis_more}
}
\vskip -0.3in
\end{figure}

We display several samples from 5 subjects in Fig.~\ref{fig:vis_more} to demonstrate the effectiveness of our MinD-3D model. Due to the varying responses of different subjects to different objects, there are some variations in the results. Overall, our model accurately reconstructs both the category and characteristics of the objects. The experiment validates the feasibility of our proposed task, Recon3DMind, and the accompanying dataset, fMRI-Shape, designed for this task.

\subsection{Visualizations of AP \& APAC Testing}
\begin{figure}[tb]
\centering
\includegraphics[width=0.9\linewidth]{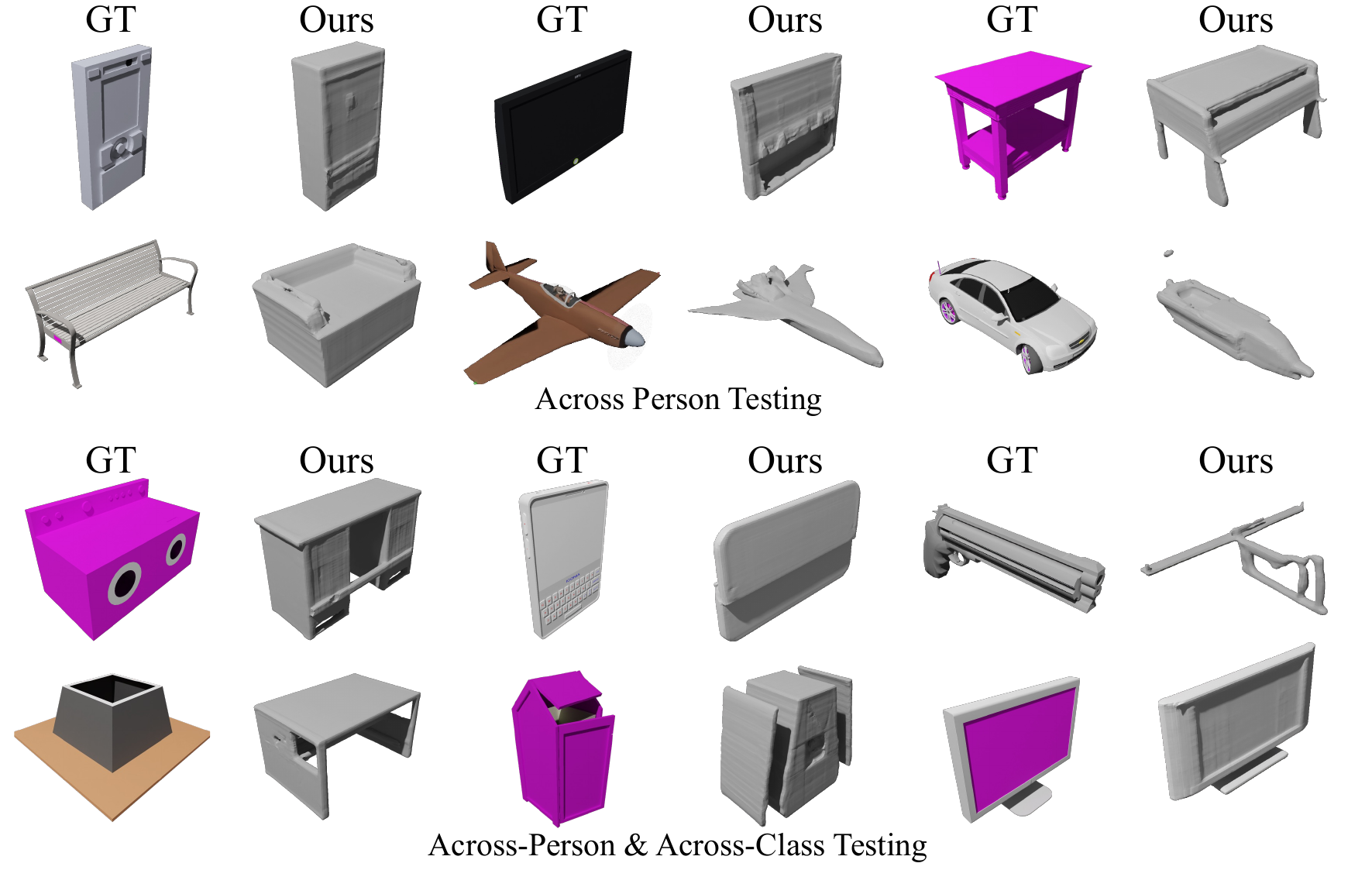}
\caption{
\textbf{Visualization of AP \& APAC testing.} AP Testing trains on Subject 1 and tests on Subject 9. APAC Testing trains on Subject 1 and tests on Subject 11.
\label{fig:vis_apac_more}
}
\end{figure}

We also showcase the reconstructed objects from AP \& APAC testing in Fig.~\ref{fig:vis_apac_more}. As highlighted in Fig.~\ref{fig:fmri}, individual differences significantly affect the outcomes. Our AP \& APAC Testing empirically confirms this. Despite the task's high difficulty, MinD-3D successfully recovers the basic shapes of the objects, establishing a baseline method for the community.

\clearpage

\subsection{Fail Cases}
\begin{figure}[h]
\centering
\includegraphics[width=0.95\linewidth]{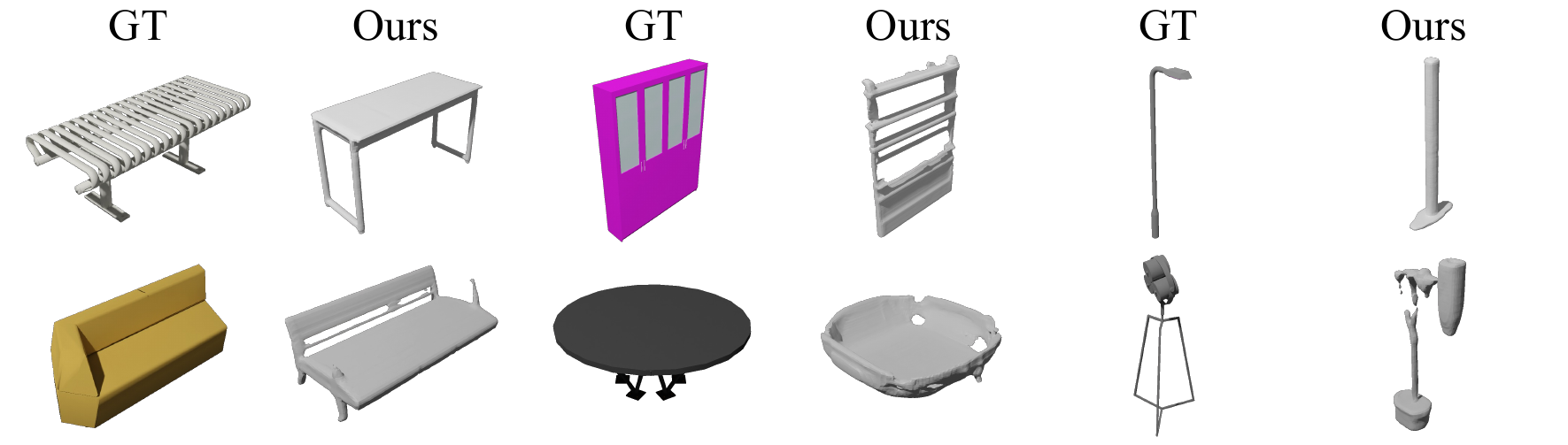}
\vskip -0.05in
\caption{
Fail cases.
\label{fig:vis_fail_cases}
}
\end{figure}

We present some failure cases in Fig.~\ref{fig:vis_fail_cases}. Although the reconstruction does not always accurately capture the entire object, it still manages to retain the shape or characteristics of the objects.

\section{More Implementation Details}
To more effectively demonstrate the details of our model and its implementation, this section introduces the necessary preliminary knowledge and provides a detailed description of our neuro-fusion encoder and latent adapted decoder.

\subsection{Neuro-Fusion Encoder}

\begin{figure}[h]
\centering
\includegraphics[width=1.0\linewidth]{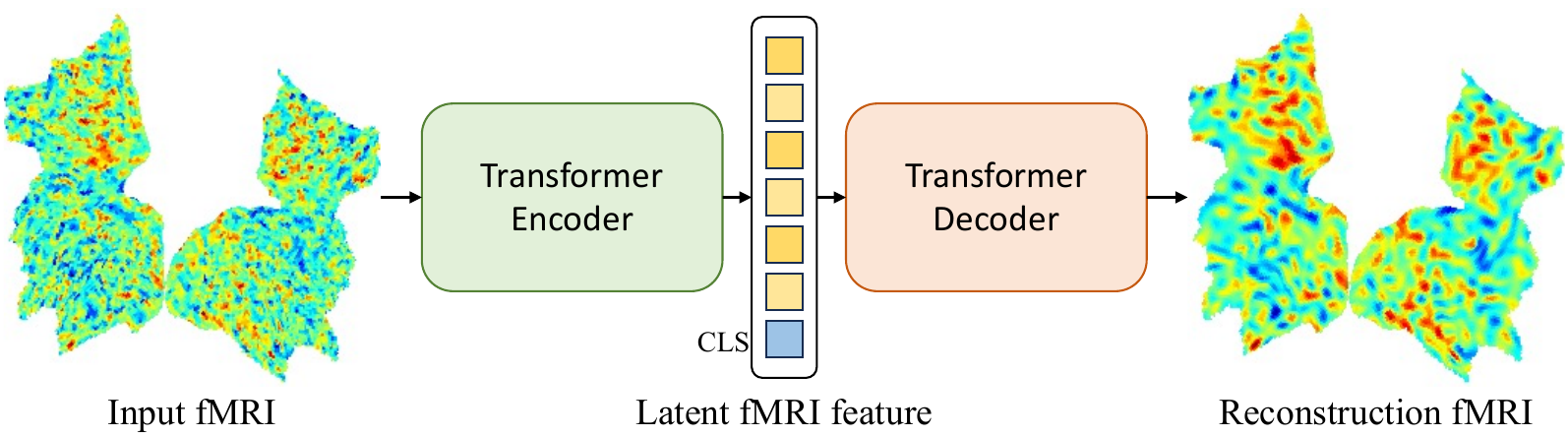}
\vskip -0.05in
\caption{
\textbf{Architecture of LEA.} During training, LEA masks all the patch-based latent codes and enables the decoder to learn to reconstruct fMRI signals from the learned [CLS] token.
}
\label{fig:lea}
\vskip -0.1in
\end{figure}

Our encoder consists of the encoder from the pre-trained LEA~\cite{qian2023semantic} and a feature aggregation module. LEA, a transformer-based auto-encoder, encodes fMRI signals and reconstructs them as depicted in Fig.~\ref{fig:lea}. Unlike the original Masked Autoencoder (MAE), which only masks some patch-based latent codes, LEA masks all such codes. This enables the decoder to learn to reconstruct fMRI signals from the learned [CLS] token during training, thereby facilitating the development of a dense fMRI representation within the [CLS] token.

LEA is trained on the UKB dataset~\cite{miller2016multimodal}. As shown in Fig.~\ref{fig:lea}, when LEA is directly applied to our fMRI-Shape dataset, it yields rough reconstructions. However, these reconstructions are not highly precise, primarily due to variations in scanners, subjects, and tasks.

Given that our input fMRI signals are in video form, to efficiently extract both semantic and structural features from this complex data, we opt to fine-tune the encoder with a small learning rate during the initial stage of training with hyperparameters provided in Tab.~\ref{tab:encoder_param}.

\begin{table}[ht] \small
\vskip -0.15in
\caption{
Hyperparameters used in neuro-fusion encoder.
\label{tab:encoder_param}
}
\vskip -0.05in
\centering
\setlength{\tabcolsep}{4mm}{
{\begin{tabular}{cccc}
\toprule
parameter  & value  & parameter  & value  \\
\midrule
patch size      & 16        & encoder depth  & 24 \\
embedding dim   & 1024      & encoder heads  & 16 \\
mask ratio      & 1.0       & fMRI frames    & 6 \\
mlp ratio       & 0.99      & image frames   & 4 \\
\bottomrule
\end{tabular}}}
\vskip -0.3in
\end{table}

\subsection{Feature Bridge Diffusion Model}
We use a transformer-based diffusion model to generate vision features. During the inference phase of the diffusion model, we concatenate the fMRI token, the time token, the predicted image token, and the learnable token. The output learnable token then serves as the predicted image token for the model's next inference. The hyperparameters of our feature bridge diffusion model are shown in Tab.~\ref{tab:diffusion_param}.

\begin{table}[ht] \small
\caption{
Hyperparameters used in feature bridge diffusion model.
\label{tab:diffusion_param}
}
\vskip -0.05in
\centering
\setlength{\tabcolsep}{4mm}{
{\begin{tabular}{cccc}
\toprule
parameter  & value  & parameter  & value  \\
\midrule
depth           & 6     & heads         & 8     \\
embedding dim   & 512   & timestamps    & 100   \\
mlp ratio       & 0.99  & token length       & 770  \\
\bottomrule
\end{tabular}}}
\vskip -0.3in
\end{table}

\subsection{Latent Adapted Decoder}

Our decoder is adapted from Argus. The original Argus model enhances the quality and diversity of generated 3D shapes by incorporating tri-plane features as latent representations and a discrete codebook for efficient quantization, leveraging the power of transformers for multi-modal conditional generation. It generates latent codes in an auto-regressive manner, similar to VQGAN, based on conditional features. Subsequently, it decodes the quantized vectors into 3D mesh. 

In our latent-adapted decoder, we add an adapter layer after every two (T=2) Transformer blocks to fuse the fMRI feature. During the second stage of training, we freeze the parameters of Argus and focus on optimizing the parameters in the adapter layers. The hyperparameters of our latent adapted decoder are detailed in Tab.~\ref{tab:decoder_param}.

\begin{table}[ht] \small
\vskip -0.1in
\caption{
Hyperparameters used in latent adapted decoder.
\label{tab:decoder_param}
}
\vskip -0.05in
\centering
\setlength{\tabcolsep}{3.5mm}{
{\begin{tabular}{cccc}
\toprule
parameter  & value  & parameter  & value  \\
\midrule
codebook size & 8192        & encoder depth  & 32 \\
codebook dim  & 512         & encoder heads  & 16 \\
embedding dim & 3072        & adapter layers  & 4 \\
mlp ratio     & 0.99        & sequence length  & 1027 \\
\bottomrule
\end{tabular}}}
\vskip -0.3in
\end{table}

\section{Features Analysis}
To understand the biological relevance of our feature extractor, in line with previous studies~\cite{wen2018neural,takagi2023high,ozcelik2023brain}, we employ a linear encoding model to project the extracted fMRI features \(c_f\) onto brain activity patterns, using subject 1 as a representative case for analysis. 

\begin{figure}[hb]
\vskip -0.1in
\centering
\includegraphics[width=0.9\linewidth]{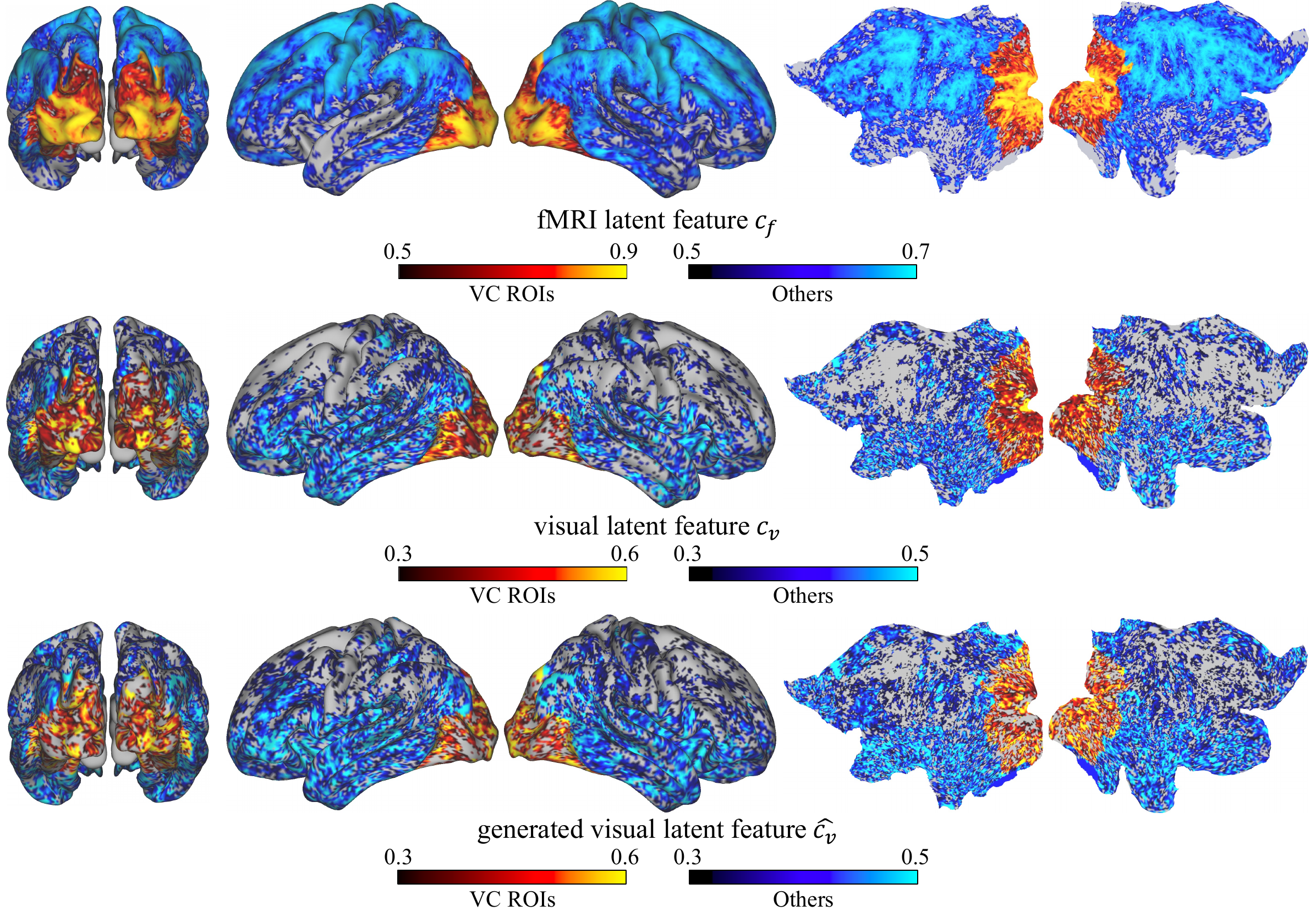}
\vskip -0.05in
\caption{
\textbf{Quality of the extracted features.} 
We map the features to brain regions via ridge regression modeling using $c_f$, $c_v$ and $\hat{c_v}$, with voxel-wise Pearson’s correlation coefficients computed between them. The visual function ROI is highlighted in red. In contrast, other regions are depicted in blue.
\label{fig:encoder}
}
\end{figure}

Specifically, we utilize ridge regression to establish the mapping between \(c_f\) and the fMRI signals within the training dataset. We then project \(c_f\) from the test set into the fMRI space and compute the Pearson correlation coefficient to assess the relationship. As illustrated in Fig.~\ref{fig:encoder}, areas positively correlated with visual Regions of Interest (ROIs) are marked in red, while other regions showing positive correlations are highlighted in blue. The results demonstrate a strong positive correlation between \(c_f\) and visual ROIs, confirming the effective extraction of relevant features. Additionally, we perform a similar correlation analysis with the visual latent feature \(c_v\), which also shows positive correlations. Notably, the correlation strength of \(c_f\) surpasses that of \(c_v\), indicating its superior relevance to brain activity. However, it is important to note that in some measures, the correlation coefficient of \(c_v\) is higher than that of \(c_f\), which highlights the complexity of the task and underscores the critical role of bridging features. Furthermore, our results indicate that our generated visual latent features achieve activation levels comparable to the ground truth (GT), emphasizing the effectiveness of our comparative learning approach and demonstrating well-aligned fMRI feature and image spaces.

\section{Future Works}
As the first attempt at the task of Reconstructing 3D Objects from the Mind using fMRI (Recon3DMind). Our proposed MinD-3D model has successfully reconstructed 3D objects that are semantically and structurally similar to their originals, thereby proving the feasibility of the task. However, our model's 3D representations, which are mesh-based, accurately outline the shape of objects but do not capture color and other details as precisely as the original objects. To improve on this, we plan to develop new methods for more effectively extracting texture and color information from fMRI signals. Our goal is to refine our reconstruction technique to produce more detailed and accurate reproductions of the original 3D objects, thereby improving the quality of the models and more faithfully capturing the originals' complexity and nuances.

Crucially, we introduce the first 3D vision fMRI dataset with contributions from 14 participants across various settings. Currently, most fMRI-based visual reconstruction methods are specific to individuals, making out-of-distribution (OOD) testing, both across individuals and classes, extremely challenging. Our proposed fMRI-Shape dataset presents an opportunity to tackle this formidable challenge. Addressing individual differences is a complex task that necessitates a collaborative effort from the research community with the goal of enhancing the adaptability of such models. Regarding our dataset, we acknowledge that relying solely on the fMRI-Shape dataset is not sufficient for comprehensively addressing the task at hand. Therefore, we plan to enrich the dataset by sampling more 3D data, including all objects from ShapeNet, and involving more participants. This effort aims to expand the fMRI-Shape dataset and further advance research in fMRI-based human brain decoding.

\end{document}